\documentclass[letterpaper, 10 pt, conference]{ieeeconf}  %

\IEEEoverridecommandlockouts                              %

\usepackage{graphics} %
\usepackage{epsfig} %
\usepackage{times} %
\usepackage{amsmath} %
\usepackage{amssymb}  %

\usepackage{bm}
\usepackage{multirow}
\usepackage{subfigure}
\usepackage{tabularx}
\usepackage{color, soul}

\newcommand\norm[1]{\left\lVert#1\right\rVert}

\definecolor{somegray}{rgb}{0.5, 0.5, 0.5}
\newcommand{\darkgrayed}[1]{\textcolor{somegray}{#1}}
\makeatletter
\newcommand*\titleheader[1]{\gdef\@titleheader{#1}}
\AtBeginDocument{%
  \let\st@red@title\@title
  \def\@title{%
    \vskip-3em
    \bgroup\normalfont\large\centering\@titleheader\par\egroup
    \vskip1.5em\st@red@title}
}
\makeatother

\titleheader{\darkgrayed{This paper has been accepted for publication at the
IEEE/RSJ International Conference on Intelligent\\ Robots and Systems (IROS), Las Vegas, 2020.
\copyright IEEE}}

\title{\LARGE \bf
Tightly-coupled Fusion of Global Positional Measurements 
\\in Optimization-based Visual-Inertial Odometry
}

\author{Giovanni Cioffi, Davide Scaramuzza%
\thanks{The authors are with the Robotics and Perception Group, Dep.  of Informatics, University of Z{\"u}rich, and Dep. of Neuroinformatics, University of Z{\"u}rich and ETH Z{\"u}rich,  Switzerland {\tt\small http://rpg.ifi.uzh.ch}.
This research was supported by the European Union’s Horizon2020 research and innovation program through the AERIAL-CORE project (H2020-2019-871479).}%
}

\begin{document}

\maketitle
\thispagestyle{empty}
\pagestyle{empty}

\begin{abstract}

Motivated by the goal of achieving robust, drift-free pose estimation in long-term autonomous navigation, in this work we propose a methodology to fuse global positional information with visual and inertial measurements in a tightly-coupled nonlinear-optimization--based estimator.
Differently from previous works, which are loosely-coupled, the use of a tightly-coupled approach allows exploiting the correlations amongst all the measurements.
A sliding window of the most recent system states is estimated by minimizing a cost function that includes visual re-projection errors, relative inertial errors, and global positional residuals. We use IMU preintegration to formulate the inertial residuals and leverage the outcome of such algorithm to efficiently compute the global position residuals.
The experimental results show that the proposed method achieves accurate and globally consistent estimates, with negligible increase of the optimization computational cost. Our method consistently outperforms the loosely-coupled fusion approach. The mean position error is reduced up to 50$\%$ with respect to the loosely-coupled approach in outdoor Unmanned Aerial Vehicle (UAV) flights, where the global position information is given by noisy GPS measurements.
To the best of our knowledge, this is the first work where global positional measurements are tightly fused in an optimization-based visual-inertial odometry algorithm, leveraging the IMU preintegration method to define the global positional factors.

\end{abstract}

\section{Introduction}

In order to achieve accurate and globally consistent pose estimates in autonomous robot navigation, different sensors are required. 
In recent years, many algorithms have been proposed, which use visual and inertial information to achieve accurate and high-rate pose estimates \cite{delmerico2018benchmark,huang2019icra}. However, such algorithms accumulate drift over time due to sensor noise and modeling errors, and are not suitable for long-term navigation.
As a consequence, global measurements are needed to achieve accurate estimates for long trajectories since their errors do not depend on the distance travelled. They can be used together with visual and inertial measurements to achieve high-rate, both locally and globally consistent estimates. 
The Global Positioning System (GPS) is an example of global position measurements widely used for localization in outdoor applications. 
However, GPS measurements are noisy and not reliable to be used as the only sensor modality for accurate localization. 
More accurate GPS systems, such as differential GPS, are possible but they require the availability of ground stations which limits the number of use cases.\\
Global position measurements were first fused with Visual-Inertial Odometry (VIO) estimates in a pose-graph optimization in~\cite{qin2019general} and~\cite{mascaro2018gomsf}. However, such systems were \textit{loosely-coupled}, meaning that the relative pose updates were estimated by the VIO algorithm independently of the global position information and only then aligned to the global frame via pose-graph optimization. 
In this way, the correlations amongst all the sensor measurements are automatically discarded resulting in sub-optimal results.

\begin{figure}[t]
\begin{center}
\includegraphics[width=1.0\linewidth]{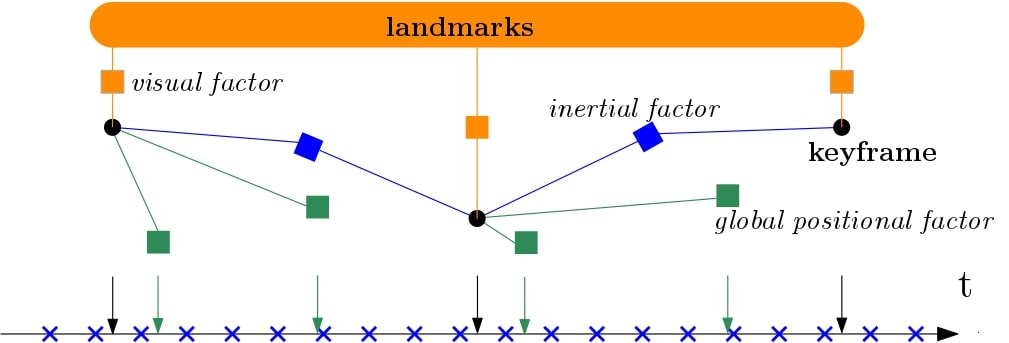}
\end{center}
   \caption{Representation of the proposed optimization-based multi-sensor fusion. We distinguish three types of factors: visual (orange), inertial (blue), and global positional factors (green). The optimization variables are the states of the keyframes in the current sliding window and the visible landmarks. In the bottom part of the figure, IMU measurements are depicted with crosses on the temporal line, while keyframes and global positional measurements are depicted with black and green arrows, respectively.}
\label{fig:eyecatcher}
\end{figure}

In this work we propose an optimization-based \textit{tightly-coupled} approach to fuse visual, inertial, and global position measurements. 
The global position measurements are used to define new factors in the optimization graph as depicted in Fig. \ref{fig:eyecatcher}. 
We define a keyframe-based sliding-window optimization as proposed in \cite{leutenegger2015keyframe}, where the main difference with respect to \cite{leutenegger2015keyframe} is the addition of the global position factors, since the number of states in the optimization does not change. 
These new error terms can be efficiently computed using the IMU preintegration algorithm \cite{lupton2011visual, forster2016manifold}.
We take advantage of the IMU preintegrated terms, already computed to define the inertial error between consecutive keyframes, to create the constraints between the position of the keyframes in the sliding window and the global position measurements. 
We show in Section \ref{sec:experiments} that, thanks to the proposed formulation of the global error terms, the increase of the computational time needed to compute and minimize the new cost function is negligible compared to the visual-inertial case.
Another important question for the multi-sensors fusion problem addressed in this work is how the number of global positional factors affects the estimates. In Section \ref{sec:experiments}, we run experiments for different numbers of global position measurements included in the estimation process.\\
In all experiments, we compare our tightly-coupled approach to a loosely-coupled based on the method proposed in~\cite{qin2019general}.
The results validate our method and show that our approach can be a step towards the target of achieving high-rate locally and globally consistent pose estimates in long-range navigation.
To the best of our knowledge, this is the first work that proposes a tightly-coupled approach to fuse global with visual and inertial measurements in an optimization-based algorithm, using the IMU preintegration method to efficiently derive the global positional error terms. 

The paper is structured as following: Section \ref{sec:related_work} contains recent work on algorithms for visual, inertial and global measurements fusion. Section \ref{sec:problem_formulation} shows our formulation of the sliding-window optimization problem. Section \ref{sec:derivation_of_global_position_residuals} introduces the IMU preintegration algorithm and shows how it can be used to derive the global positional residuals.
Section \ref{sec:experiments} contains experiments and discussions. Section \ref{sec:conclusion} concludes the paper.

\subsection{Related Work}\label{sec:related_work}

Two major approaches can be found in the literature to address the visual, inertial, and global position fusion problem: \textit{filtering methods} and \textit{smoothing methods}.

\textit{Filtering methods}: Filtering methods carry out efficient estimation by only updating the latest state. 
Many filter-based approaches involving visual and inertial measurements are inspired by the work in~\cite{mourikis2007multi}, where an Extended Kalman Filter (EKF) was proposed to perform visual-inertial odometry. 
In~\cite{weiss2012versatile}, an EKF was proposed to fuse inertial data, GPS measurements and vision-based pose estimates. 
In this case, the poses estimated by an independent (i.e., loosely-coupled) visual odometry algorithm were fused with inertial and GPS measurements in a subsequent estimation step.
In~\cite{Lee2020ICRA}, the EKF includes online calibration of IMU-GPS extrinsics and time offset.

\textit{Smoothing methods}: Smoothing algorithms are classified as \textit{full-} or \textit{fixed-lag} smoothers.
\textit{Full-lag} smoothers estimate the complete history of the states. 
They guarantee the highest accuracy but incur high computational cost.
In~\cite{indelman2013information}, it was proposed to use incremental smoothing technique~\cite{kaess2012isam2} and IMU preintegration to reduce the computational cost of the full-batch optimization.
In~\cite{cucci2017bundle}, achieving high accuracy was prioritized over an online implementation.
This work was subsequently extended in~\cite{cucci2019raw} to include an extended version of the IMU preintegration algorithm, incorporating gravity and Earth rotation in the IMU model.
\textit{Fixed-lag} smoothers (or sliding window estimators) estimate a window of the latest states while marginalizing out the previous states \cite{leutenegger2015keyframe}. 
This approach is more computational efficient than \textit{full-lag} smoothers but less accurate due to accumulation of linearization errors in the marginalization \cite{hesch2014camera}.\\
In~\cite{qin2019general}, the global position measurements were fused with poses estimated by a VIO algorithm in a sliding window pose-graph optimization of the most recent robot states. Similarly in~\cite{mascaro2018gomsf}, an independent VIO algorithm provided pose estimates that were successively fused with GPS measurements in a pose-graph optimization. Differently from~\cite{qin2019general}, in~\cite{mascaro2018gomsf} the pose-graph contains an additional node representing the origin of the local coordinate frame in order to constrain the absolute orientation. However, both these approaches were loosely-coupled, i.e. the relative pose estimates were provided by an independent VIO algorithm.
Differently from~\cite{qin2019general, mascaro2018gomsf}, we propose a tightly-coupled approach where all the measurements are included in a common optimization problem thus considering the correlations amongst them. In~\cite{leutenegger2015keyframe}, it was shown that for the visual-inertial case considering all measurement correlations is crucial for high precision estimates.
In~\cite{yu2019gps}, it was proposed a tightly-coupled sliding window optimization for visual and inertial measurements with loosely-coupled GPS refinement. The GPS measurements were assumed to be available at low-rate and they were given the same time stamp of the temporally closest image in order to be included in the sliding window. Differently from~\cite{yu2019gps}, we tightly couple the global position measurements using the IMU preintegration algorithm to efficiently derive the global positional factors. This allows to add multiple global factors per keyframe in the sliding window with negligible extra computational cost.

\section{Problem Formulation}\label{sec:problem_formulation}

\subsection{Notation}

\begin{figure}[t]
\begin{center}
\includegraphics[width=0.9\linewidth]{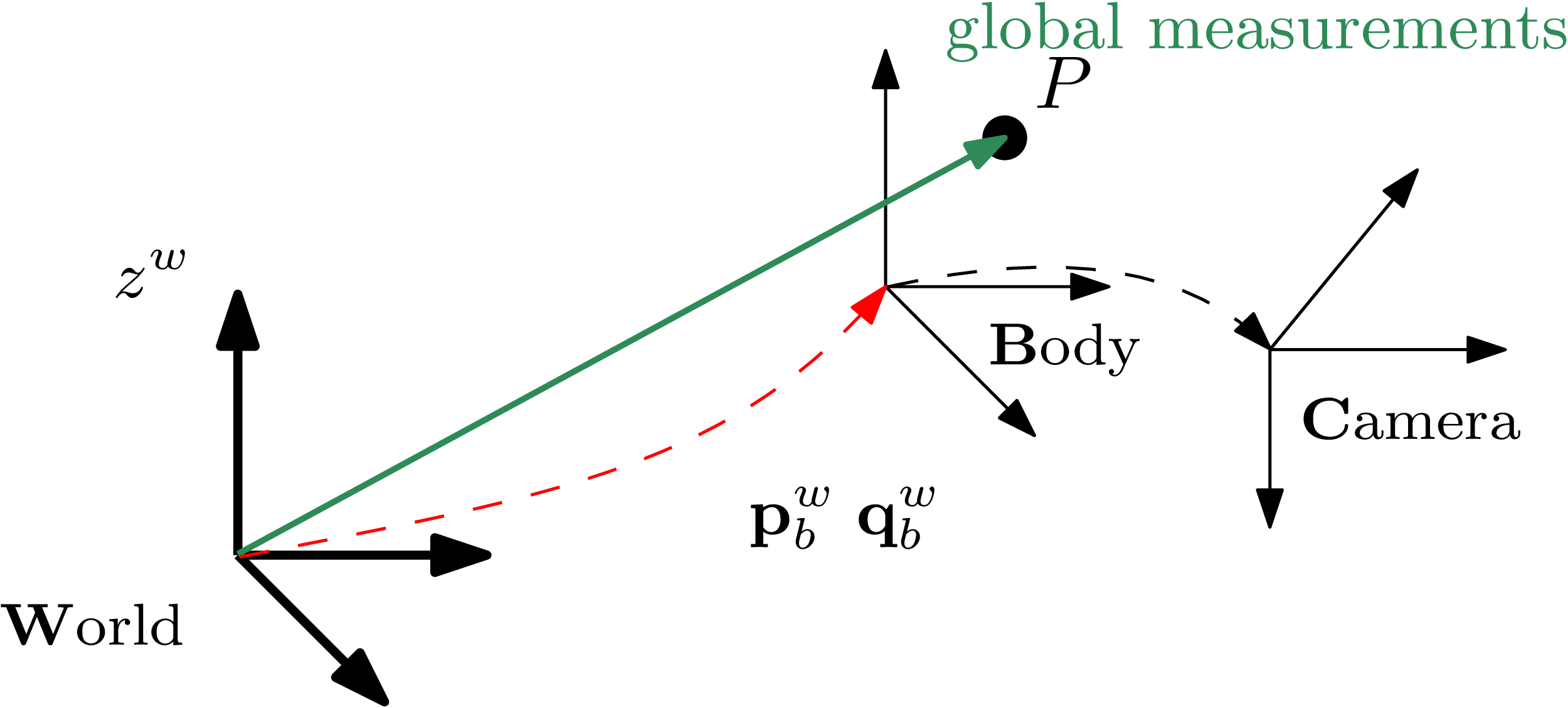}
\end{center}
   \caption{Reference frames used in this work. }
\label{fig:reference_frames}
\end{figure}

The coordinate frames used are depicted in Fig. \ref{fig:reference_frames}.
$W$ represents the world frame. We assume the direction of the gravity aligned to $z^{w}$ axis. $B$ is the body frame and corresponds to the IMU frame. The camera frame is denoted by $C$. We use the notation $(\cdot)^{w}$ to represent a quantity in the world frame $W$. Similar notation applies for every reference frame.
We use $\mathbf{p}_{b_{k}}^{w}$ and $\mathbf{q}_{b_{k}}^{w}$ to represent the position and orientation of $B$ with respect to $W$ at time $t_k$.
The rotation matrix representation is $\mathbf{R}_{b_k}^{w}$. 
The velocity of $B$ expressed in $W$ at time $t_k$ is $\mathbf{v}_{b_{k}}^{w}$. 
Global position measurements are given by $\mathbf{p}_{p_{k}}^{w}$, where $P$ is a point rigidly attached to $B$ by $\mathbf{p}_{p}^{b}$. 
For example, the point $P$ could represent the position of the receiver antenna in the case of GPS measurements. 
The value of $\mathbf{p}_{p}^{b}$ can be obtained from the calibration of the system.
The notation $\hat{(\cdot)}$ is used to represent noisy measurements.

The keyframe-based sliding window optimization variables are $\mathcal{X} = \{ \mathcal{L}, \mathcal{X}_B \}$,
where $\mathcal{L}$ comprises the position of the 3D landmarks visible in the sliding window and $\mathcal{X}_B = [\mathbf{x}_1, \cdots, \mathbf{x}_K]$ comprises the system states,
with $K$ the total number of keyframes in the sliding window.
The system state $\mathbf{x}_{k}$ at time $t_k$ is given by the body position $\mathbf{p}_{b_k}^{w}$, the body orientation quaternion $\mathbf{q}_{b_k}^{w}$, the body velocity $\mathbf{v}_{b_k}^{w}$, accelerometer $\mathbf{b}_{a_k}$, and gyroscope biases $\mathbf{b}_{g_k}$: $\mathbf{x}_{k} = [\mathbf{p}_{b_k}^{w}, \mathbf{q}_{b_k}^{w}, \mathbf{v}_{b_k}^{w}, \mathbf{b}_{a_k}, \mathbf{b}_{g_k}]$. 

\subsection{Optimization-based Visual, Inertial, and Global Information Fusion}

The keyframe-based visual-inertial localization and mapping problem is formulated as a joint nonlinear optimization which solves for the maximum a posteriori estimate of $\mathcal{X}$. Using the problem formulation as proposed in \cite{leutenegger2015keyframe} with some minor changes, the cost function to minimize is written as
\begin{equation}\label{eq:J_vio}
    J_{VI}(\mathcal{X}) =  \sum_{k = 0}^{K-1} \sum_{j \in \mathcal{J}_k} \norm{ \mathbf{e}_{\mathbf{v}}^{j,k} }_{\mathbf{W}_{\mathbf{v}}^{j,k}}^{2} +  \sum_{k = 0}^{K-1} \norm{ \mathbf{e}_{\mathbf{i}}^{k} }_{\mathbf{W}_{\mathbf{i}}^{k}}^{2} + \norm{ \mathbf{e}_{\mathbf{p}} }^{2}.
\end{equation}
$J_{VI}(\mathcal{X})$ contains the weighted visual $\mathbf{e}_{\mathbf{v}}$, inertial $\mathbf{e}_{\mathbf{i}}$, and marginalization residuals $\mathbf{e}_{\mathbf{p}}$.\\
The visual residuals are $\mathbf{e}_{\mathbf{v}}^{j,k} = \mathbf{z}^{j,k} - h(\mathbf{l}_{j}^{w})$, 
which describe the re-projection error of the landmark $\mathbf{l}_{j}^{w} \in \mathcal{J}_k$, where $\mathcal{J}_k$ is the set containing all the visible landmarks from the keyframe $k$ in the sliding window. The function $h(\cdot)$ denotes the camera projection model and $\mathbf{z}^{j,k}$ the 2D image measurement. We refer to \cite{leutenegger2015keyframe} for additional details.
The inertial residuals $\mathbf{e}_{\mathbf{i}}$ are formulated using the IMU preintegration algorithm as proposed in \cite{forster2016manifold}, \cite{qin2018vins}.
The derivation of the global positional residuals is inspired by the IMU preintegration algorithm as we describe in Section \ref{sec:derivation_of_global_position_residuals}.
The error term $\mathbf{e}_{\mathbf{p}}$ denotes the prior information obtained from marginalization. 
We adopt the marginalization strategy proposed in \cite{leutenegger2015keyframe}. Namely, when a new frame is inserted in the sliding window we distinguish two cases.
In the case the oldest state in the sliding window is not a keyframe, it is marginalized out and all its landmarks are dropped to keep sparsity.
In the case the oldest state is a keyframe, the landmarks visible from such frame but not in the most recent keyframe are also marginalized out.

Global positional residuals are added to (\ref{eq:J_vio}) to derive the cost function proposed in this work, as
\begin{equation}\label{eq:J}
    J(\mathcal{X}) = J_{VI}(\mathcal{X}) + \sum_{k = 0}^{K-1} \sum_{j \in \mathcal{G}_{k}} \norm{ \mathbf{e}_{\mathbf{g}}^{j,k} }_{\mathbf{W}_{\mathbf{g}}^{k}}^{2},
\end{equation}
where $\mathcal{G}_{k}$ contains the global positional measurements connected to the state $\mathbf{x}_{k}$ by an error term.\\
Next, we derive the global residual terms $\mathbf{e}_{\mathbf{g}}^{j,k}$ leveraging the outcome of the IMU preintegration algorithm and infer the residual weights $\mathbf{W}_{\mathbf{g}}^{k}$. 

\section{Derivation of Global Position Residuals}\label{sec:derivation_of_global_position_residuals}

\subsection{IMU Preintegration}\label{sec:imu_preintegration}
In this section, we review the IMU preintegration algorithm focusing on the derivation of the quantities then utilized in Section \ref{sec:global_position_residuals} for the formulation of the global positional residuals. We use the IMU preintegration derivation proposed in \cite{qin2018vins}, which is based on the continuous-time quaternion-based formulation in \cite{shen2015tightly} and includes the manipulation of IMU biases as in \cite{forster2016manifold}.\\
IMU residuals are formulated as relative constraints between consecutive states using accelerometer $\hat{\mathbf{a}}_t = \mathbf{a}_t + \mathbf{b}_{a_{t}} + \mathbf{R}_{w}^{t}\mathbf{g}^{w} + \mathbf{n}_{a}$, and gyroscope $\hat{\mathbf{w}}_t = \mathbf{w}_t + \mathbf{b}_{w_{t}} + \mathbf{n}_{w}$ measurements. The accelerometer and gyroscope additive noises are modeled as additive Gaussian noise $\mathbf{n}_{a} \thicksim \mathcal{N}(\mathbf{0}, \sigma_{a}^2 \cdot \mathbf{I})$ and $\mathbf{n}_{w} \thicksim \mathcal{N}(\mathbf{0}, \sigma_{w}^2 \cdot \mathbf{I})$, where $\mathbf{I}$ is the identity matrix. 
Biases are modeled as random walks $\mathbf{\Dot{b}}_{a_{t}} = \bm{\eta}_{b_{a}}$ and $\mathbf{\Dot{b}}_{w_{t}} = \bm{\eta}_{b_{w}}$, with $\bm{\eta}_{b_{a}} \thicksim \mathcal{N}(\mathbf{0}, \sigma_{b_{a}}^2 \cdot \mathbf{I})$ and $\bm{\eta}_{b_{w}} \thicksim \mathcal{N}(\mathbf{0}, \sigma_{b_{w}}^2 \cdot \mathbf{I})$.\\
Given the time interval $[t_k, t_{k+1}]$, $\mathbf{p}_{b_{k}}^{w}$, $\mathbf{v}_{b_{k}}^{w}$, and $\mathbf{q}_{b_{k}}^{w}$ can be propagated in such time interval by using the accelerometer and gyroscope measurements.
The propagation in the world frame requires the knowledge of the initial state.
This implies that every time the estimate of the initial state changes, e.g. when it is updated in an optimization step, repropagation is needed. 
The main benefit of the IMU preintegration algorithm is to avoid the need of repropagation at every optimization step, which results in saving of valuable computational resources.\\
The propagation is executed in the local frame $B_k$ instead of the world frame as
\begin{align}
    \mathbf{R}_{w}^{b_k} \mathbf{p}_{b_{k+1}}^{w} & = \mathbf{R}_{w}^{b_w} (\mathbf{p}_{b_{k}}^{w} + \mathbf{v}^{w}_{b_{k}} \Delta t_{k} - \frac{1}{2} \mathbf{g}^{w} \Delta t_{k}^{2} ) + \bm{\alpha}_{b_{k+1}}^{b_{k}} \nonumber \\
    \mathbf{R}_{w}^{b_k} \mathbf{v}_{b_{k+1}}^{w} & = \mathbf{R}_{w}^{b_k} (\mathbf{v}_{b_{k}}^{w} - \mathbf{g}^{w} \Delta t_{k}) + \bm{\beta}_{b_{k+1}}^{b_{k}} \nonumber \\
    \mathbf{q}^{b_{k}}_{w} \otimes \mathbf{q}_{b_{k+1}}^{w} & = \bm{\gamma}_{b_{k+1}}^{b_{k}}, \label{eq:state_propagation_local}
\end{align}
where $\bm{\alpha}_{b_{k+1}}^{b_{k}}$, $\bm{\beta}_{b_{k+1}}^{b_{k}}$, and $\bm{\gamma}_{b_{k+1}}^{b_{k}}$ are the preintegration terms, which only depend on the inertial measurements as well as biases.\\    
In the discrete-time case using Euler numerical integration method, the mean of $\bm{\alpha}$, $\bm{\beta}$ and $\bm{\gamma}$ can be computed recursively as
\begin{align}
    \hat{\bm{\alpha}}_{i+1}^{b_{k}} & = \hat{\bm{\alpha}}_{i}^{b_{k}} + \hat{\bm{\beta}}_{i}^{b_{k}} \delta t + \frac{1}{2}\mathbf{R}(\hat{\bm{\gamma}}_{i}^{b_{k}})(\hat{\mathbf{a}}_{i} - \mathbf{b}_{a_{i}})\delta t^2 \nonumber \\
    \hat{\bm{\beta}}_{i+1}^{b_{k}} & = \hat{\bm{\beta}}_{i}^{b_{k}} + \mathbf{R}(\hat{\bm{\gamma}}_{i}^{b_{k}})(\hat{\mathbf{a}}_{i} - \mathbf{b}_{a_{i}})\delta t \nonumber \\
    \hat{\bm{\gamma}}_{i+1}^{b_{k}} & = \hat{\bm{\gamma}}_{i}^{b_{k}} \otimes \begin{bmatrix} 1 \\ \frac{1}{2}(\hat{\bm{w}}_{i} - \mathbf{b}_{w_{i}})\delta t \end{bmatrix},
    \label{eq:preintegration_terms_recursive_formulation}
\end{align}
where $\bm{\alpha}_{b_{k}}^{b_{k}} = \bm{\beta}_{b_{k}}^{b_{k}}$ = 0 and $\bm{\gamma}_{b_{k}}^{b_{k}}$ is equal to the identity quaternion. $\mathbf{R}(\hat{\bm{\gamma}}_{j}^{k})$ is the rotation matrix representation of $\hat{\bm{\gamma}}_{j}^{k}$.
The covariance $\mathbf{P}_{b_{k+1}}^{b_{k}}$ can be also calculated recursively, we refer to \cite{qin2018vins} for the derivation.\\
As proposed in \cite{forster2016manifold}, we update $\bm{\alpha}_{b_{k+1}}^{b_{k}}, \bm{\beta}_{b_{k+1}}^{b_{k}}, \bm{\gamma}_{b_{k+1}}^{b_{k}}$ using their first-order approximation with respect to the biases if the change in the estimate of the biases is small. 
Otherwise, propagation is redone.
The inertial residuals $\mathbf{e}_{\mathbf{i}}^{k}$ are derived from (\ref{eq:state_propagation_local}) as
\begin{equation}\label{eq:imu_position_residual}
    \mathbf{e}_{\mathbf{i}}^{k} = \begin{bmatrix} \mathbf{R}_{w}^{b_w} (\mathbf{p}_{b_{k+1}}^{w} - \mathbf{p}_{b_{k}}^{w} - \mathbf{v}^{w}_{b_{k}} \Delta t_{k} + \frac{1}{2} \mathbf{g}^{w} \Delta t_{k}^{2} ) - \hat{\bm{\alpha}}_{b_{k+1}}^{b_{k}} \\ \mathbf{R}_{w}^{b_w} (\mathbf{v}_{b_{k+1}}^{w} - \mathbf{v}_{b_{k}}^{w} + \mathbf{g}^{w} \Delta t_{k} ) - \hat{\bm{\beta}}_{b_{k+1}}^{b_{k}} \\ 2[(\mathbf{q}_{b_{k}}^{w})^{-1} \otimes \mathbf{q}_{b_{k+1}}^{w} \otimes (\hat{\bm{\gamma}}_{b_{k+1}}^{b_{k}})^{-1}]_{xyz} \\ \mathbf{b}_{a_{k+1}} - \mathbf{b}_{a_{k}} \\ \mathbf{b}_{w_{k+1}} - \mathbf{b}_{w_{k}} \end{bmatrix}.
\end{equation}
We leverage the formulation of the position error term in (\ref{eq:imu_position_residual}) to derive the global positional residuals as formulated in the next section. 

\subsection{Global Position Residuals}\label{sec:global_position_residuals}
The global positional measurements are given by $\{\mathbf{p}_{p_j}^{w}\}$ at time $\{t_j\}$. We model the measurement uncertainty with additive Gaussian noise so that
\begin{equation}\label{eq:global_measurements_uncertainty}
    \hat{\mathbf{p}}_{p_j}^{w} = \mathbf{p}_{p_j}^{w} + \mathbf{n}_p,
\end{equation}
where $\mathbf{n}_p\thicksim\mathcal{N}(\mathbf{0}, \sigma_{p}^2 \cdot \mathbf{I})$.
Given a state in the current sliding window $\mathbf{x}_{k}$ at time $t_k$ and a measurement $\hat{\mathbf{p}}_{p_j}^{w}$ at time $t_j \in [t_k, t_{k+1})$ the global position residual is defined as
\begin{equation}\label{eq:global_position_residual}
    \mathbf{e}_{\mathbf{g}}^{j,k} = \mathbf{R}_{w}^{b_k} (\hat{\mathbf{p}}_{b_{j}}^{w} - \mathbf{p}_{b_{k}}^{w} - \mathbf{v}^{w}_{b_{k}} \Delta t_{k} + \frac{1}{2} \mathbf{g}^{w} \Delta t_{k}^{2} ) - \hat{\bm{\alpha}}_{b_{j}}^{b_{k}},
\end{equation}
where the measurement $\hat{\mathbf{p}}_{p_j}^{w}$ is transformed in $\hat{\mathbf{p}}_{b_{j}}^{w}$ as
\begin{equation}\label{eq:pj_bj_transformation}
    \hat{\mathbf{p}}_{b_{j}}^{w} = \hat{\mathbf{p}}_{p_{j}}^{w} - \mathbf{R}_{b_{j}}^{w} \mathbf{p}_{p}^{b},
\end{equation}
with $\mathbf{R}_{b_{j}}^{w} = \mathbf{R}_{b_{k}}^{w} \hat{\bm{\gamma}}_{j}^{k}$.\\
To define the global residuals, the state position is propagated using inertial measurements in the time interval $[t_k, t_j]$. We express the error term in (\ref{eq:global_position_residual}) in the reference frame $B_{k}$ and take advantage of the computation of the preintegration terms in (\ref{eq:preintegration_terms_recursive_formulation}). In fact, $\hat{\bm{\alpha}}_{b_{j}}^{b_{k}}$ can be efficiently obtained during the recursive calculation of $\hat{\bm{\alpha}}_{b_{k+1}}^{b_{k}}$, in (\ref{eq:preintegration_terms_recursive_formulation}), since $t_j < t_{k+1}$ and imu measurements are buffered in $[t_k, t_{k+1}]$. The same applies for $\hat{\bm{\gamma}}_{j}^{k}$. 
This allows to minimize the computational time required to include the error term (\ref{eq:global_position_residual}) in the cost function (\ref{eq:J}).       
To derive the residual weights $\mathbf{W}_{\mathbf{g}}^{k}$, we rewrite (\ref{eq:global_position_residual}) as
\begin{align}\label{eq:global_position_residual_rewritten}
    \mathbf{e}_{\mathbf{g}}^{j,k} = & \mathbf{R}_{w}^{b_k} (-\mathbf{p}_{b_{k}}^{w} - \mathbf{v}^{w}_{b_{k}} \Delta t_{k} + \frac{1}{2} \mathbf{g}^{w} \Delta t_{k}^{2} )  \nonumber \\& - \hat{\bm{\alpha}}_{b_{j}}^{b_{k}} + \mathbf{R}_{w}^{b_k} \hat{\mathbf{p}}_{p_{j}}^{w} 
    - \mathbf{R}(\hat{\bm{\gamma}}_{j}^{k}) \mathbf{p}_{p}^{b}.
\end{align}
In (\ref{eq:global_position_residual_rewritten}), $\hat{\bm{\alpha}}_{b_{j}}^{b_{k}}$, $\hat{\mathbf{p}}_{p_{j}}^{w}$ and $\hat{\bm{\gamma}}_{j}^{k}$ are the noisy measurements. 
The covariance of $\hat{\bm{\gamma}}_{j}^{k}$ depends on gyroscope noise and bias. Since gyroscope noise is already considered in the computation of $\hat{\bm{\alpha}}_{b_{j}}^{b_{k}}$ (the reader can refer to \cite{qin2018vins} and \cite{forster2016manifold} for additional details) and it is usually smaller than accelerometer noise, we omit $\hat{\bm{\gamma}}_{j}^{k}$ in the derivation of $\mathbf{W}_{\mathbf{g}}^{k}$.\\
As consequence, the residual weights depend on the covariance of $\hat{\bm{\alpha}}_{b_{j}}^{b_{k}}$ and $\hat{\mathbf{p}}_{p_{j}}^{w}$ as
\begin{equation}\label{eq:global_residuals_weights}
    \mathbf{W}_{\mathbf{g}}^{k} = _{\hat{\bm{\alpha}}}\mathbf{P}_{b_{j}}^{b_{k}} + \mathbf{R}_{w}^{b_k} (\sigma_{p}^2 \cdot \mathbf{I}) (\mathbf{R}_{w}^{{b_k}})^t ,
\end{equation}
where $_{\hat{\bm{\alpha}}}\mathbf{P}_{b_{j}}^{b_{k}}$ is the top-left 3x3 part of $\mathbf{P}_{b_{j}}^{b_{k}}$.
The covariance $\mathbf{P}_{b_{j}}^{b_{k}}$ differs from the covariance of the inertial residuals (i.e., $\mathbf{P}_{b_{k+1}}^{b_{k}}$) since it is derived from a sub-set of the inertial measurements in $[t_k, t_{k+1}]$.\\
When a state connected to global position residuals needs to be marginalized, the global residuals are transformed in the prior linear error term together with inertial and visual residuals.

\textbf{Sampling Strategy:} We define $\mathcal{G}_{k}$ as the set containing the global position measurements in the time interval $[t_k, t_{k+1})$, which are connected to the state $\mathbf{x}_{k}$ by error term in~(\ref{eq:global_position_residual}). 
N is the cardinality of $\mathcal{G}_{k}$ such that N = $|\mathcal{G}_{k}|$.
Since we use the recursive formulation of the IMU preintegrated terms in~(\ref{eq:preintegration_terms_recursive_formulation}) to compute $\hat{\alpha}_{b_{j}}^{b_{k}}$ in (\ref{eq:global_position_residual}), increasing N only has a minor affect on the optimization computational cost.
As soon as a new measurement is available, it is included in $\mathcal{G}_{k}$ and a new residual (\ref{eq:global_position_residual}) is added to the optimization. 
In Section \ref{sec:experiments}, we evaluate how different maximum values of N affect the pose estimates. 

\section{Experiments}\label{sec:experiments}
We evaluated our approach on two visual-inertial datasets with global position measurements: 
an indoor one (the EuRoC dataset~\cite{burri2016euroc}, Section~\ref{sec:EuroC}) and an outdoor one (from~\cite{mascaro2018gomsf}, Section~\ref{UAVexperiment}).
Since the EuRoC dataset provides global position measurements from a motion capture system, we corrupted the motion capture system measurements with Gaussian noise to simulate noisy global position measurements. The second dataset, instead, provides global position measurements from a GPS and ground-truth from a total station.

As a vision front-end, we used the one of SVO~\cite{forster2016svo}. 
The vision front-end deals with feature detection and tracking from images. 
Features correctly tracked in subsequent frames are then triangulated and added to the sliding-window optimization. 
The vision front-end is also responsible for the selection of the keyframes.
We limited the number of keyframes in the sliding-window to 10.
We used the Ceres Solver~\cite{ceres-solver} to solve the optimization problem.
The vision front-end and the sliding-window solver run in two separate threads.
All the experiments ran on a laptop equipped with a 2.60GHz Intel Core i7 CPU.
\subsection{EuRoC Dataset}\label{sec:EuroC}
\subsubsection{Setup}
The EuRoC dataset contains eleven sequences recorded from a hex-rotor helicopter. Five sequences are recorded in an industrial machine hall and six in an office room.
The sequences recorded in the industrial hall are labeled as MH\_ and those in the office room as V\_. Every sequence is classified as easy, medium, or hard depending on illumination conditions, scene texture and vehicle motion. Hard sequences contain challenging illumination conditions and fast motion.
Hardware synchronized stereo images and IMU measurements are available at a rate of 20 Hz and 200 Hz, respectively.
We ran the experiments in a monocular setup using only images from the left camera.
In every sequence, a motion capture system was used to record ground-truth.
The ground-truth measurements were corrupted with zero-mean Gaussian noise to simulate noisy global position measurements.
The Gaussian noise was defined as $\mathbf{n}_{mc} \thicksim \mathcal{N}(\mathbf{0}, \sigma_{mc}^{2} \cdot \mathbf{I})$, $\sigma_{mc}$ = $20$ cm.
The additive Gaussian noise $\mathbf{n}_p$ in (\ref{eq:global_measurements_uncertainty}) was set equal to $\mathbf{n}_{mc}$.
For initialization, we set the initial position equal to the corresponding noisy motion capture system measurement. 
\subsubsection{Results}
The proposed method was evaluated in terms of estimated trajectory accuracy and solver time. We used the trajectory evaluation toolbox in \cite{Zhang18iros} to compute the evaluation metrics.
Each state $\mathbf{x}_{k}$ in the optimization window is connected to N = $|\mathcal{G}_{k}|$ global positional residuals, where $\mathcal{G}_{k}$ is the set containing the global positional measurements in the time interval $[t_k, t_{k+1})$. 
We were interested in how the cardinality of $\mathcal{G}_{k}$ influences the trajectory estimates and for the experiments in this section we evaluated $|\mathcal{G}_{k}|$ = [1, 2, 3, 4]. 
We also included as reference the VIO estimates, in this case no global residual terms are included in the optimization and the sliding window cost function is (\ref{eq:J_vio}).

\begin{figure}[t]
\begin{center}
\includegraphics[width=1.0\linewidth,height=1.0\linewidth]{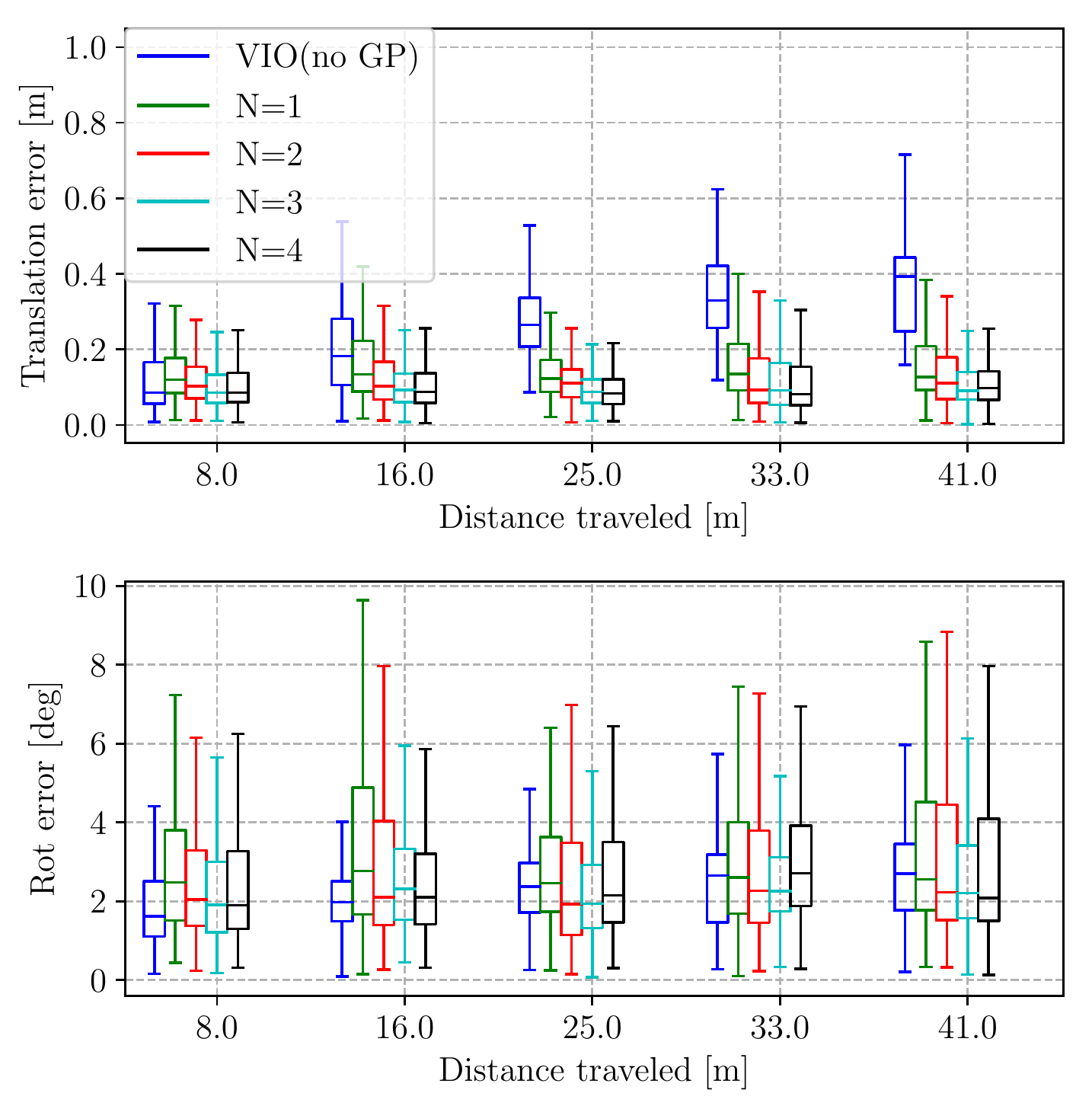}
\end{center}
   \caption{Relative translation and rotation error in EuRoC V203 difficult. Each plot contains evaluation for different values of N = $|\mathcal{G}_{k}|$ as well as VIO-only estimates, i.e. global positional (GP) measurements are not included in the estimation process.}
\label{fig:pose-error-V203}
\end{figure}

\begin{figure}[t]
\begin{center}
\includegraphics[width=1.0\linewidth]{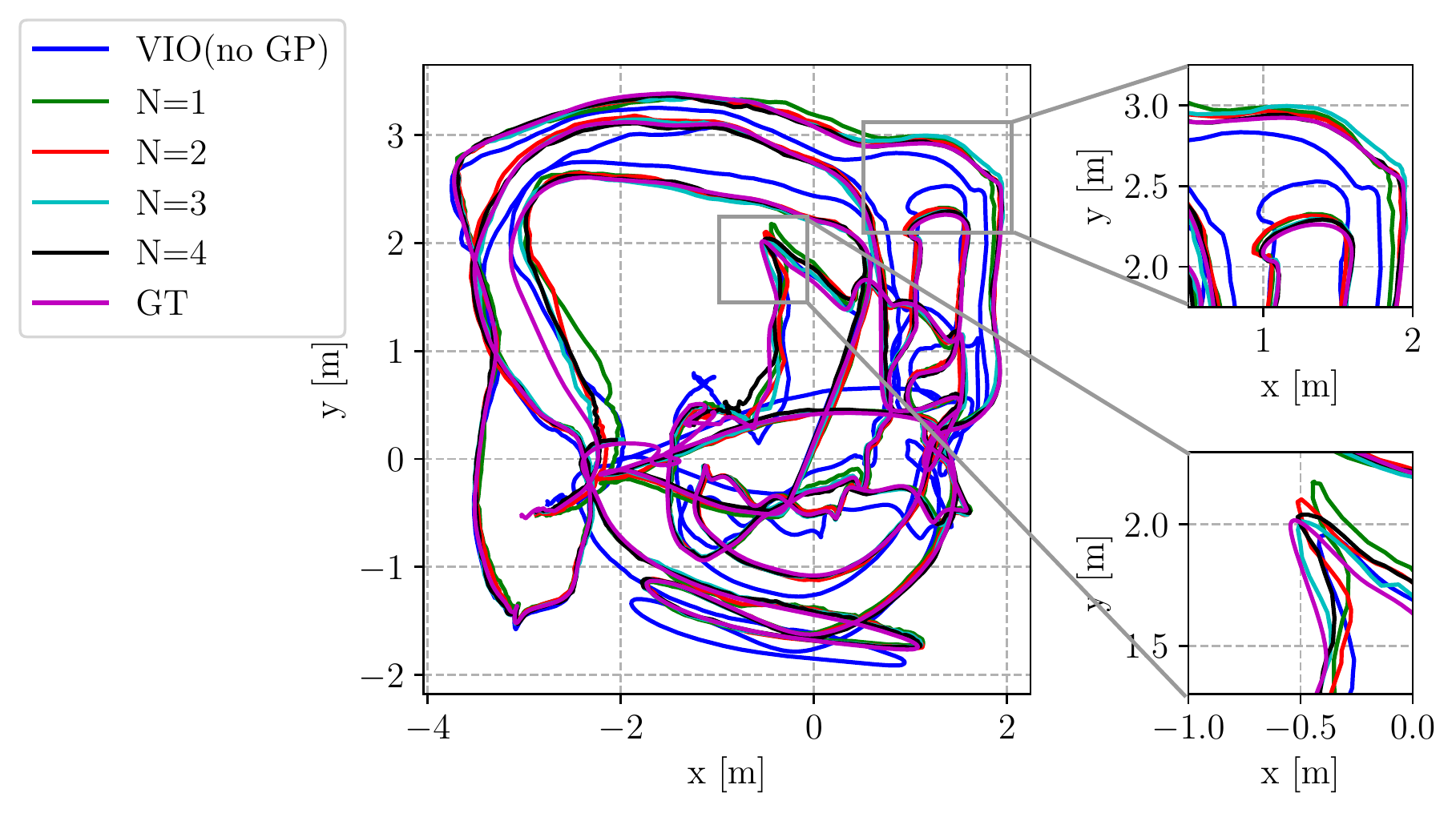}
\end{center}
   \caption{Trajectory top-view of the sequence V203 difficult. The two zoomed-in sections in the right column highlight how the drift accumulated in the VIO-only case is corrected depending on the number of global residuals N per keyframe included in the sliding window.}
\label{fig:trajectory-V203}
\end{figure}

\textbf{Accuracy:} Table \ref{table:rmse_euroc} contains the absolute trajectory error (ATE)~\cite{Zhang18iros,sturm2012benchmark} obtained on all the EuRoC sequences.
We included the results for the VIO-only case, i.e. without fusion of global positional (GP) measurements, the loosely-coupled approach, and our proposed tightly-coupled approach with N $\in$ [1,2,3,4].
For the VIO-only case, the estimated trajectory is aligned to the ground-truth using the \textit{posyaw} alignment method in \cite{Zhang18iros}. When N $\geq$ 1 (i.e., when GP measurements are considered), no alignment is applied.
Each configuration was run three times and the ATE median value is reported.
By comparing the VIO-only to the loosely- and tightly-coupled results, we can see that the ATE decreases when global positional factors are included in the sliding window.
In our proposed tightly-coupled approach, the residual in (\ref{eq:global_position_residual}) constrains the position estimate at $t_k$ to be consistent with the global positional measurement allowing to reduce the error that accumulates in the visual-inertial estimates.
The largest improvement between the VIO-only case and the tightly-coupled approach with N=1 is in sequence MH05, where the ATE decreases from 0.306 m to 0.056 m.
The estimate accuracy is improved by adding more global residuals per keyframe (i.e., N$>$1).
Increasing N from 1 to 2 allows to reduce the ATE in every sequence, with the largest and smallest improvements in sequence V203 and MH01, respectively. The average decrease of the ATE for all the EuRoC sequences is equal to 0.010 m.
Increasing N from 2 to 3 improves the ATE in every sequence but the benefit is smaller than the previous case (i.e., increasing N from 1 to 2). The average decrease of the ATE on all the EuRoC sequences is equal to 0.005 m.
A further increase of N from 3 to 4 has a minor effect on the estimate accuracy. The ATE remains the same for the sequence MH01, MH05, V101, V201, and V203, and it slightly decreases for other sequences.\\
\begin{table}[]
\caption{ATE [m] on the EuRoC sequences.}
\begin{tabular}{lcccccc}
\hline
\multirow{2}{*}{\textbf{Sequence}} &
  \multirow{2}{*}{\textbf{\begin{tabular}[c]{@{}c@{}}VIO-only\\ (no GP)\end{tabular}}} &
  \multirow{2}{*}{\textbf{\begin{tabular}[c]{@{}c@{}}Loosely-\\ coupled\\ (\cite{qin2019general})\end{tabular}}} &
  \multicolumn{4}{c}{\textbf{\begin{tabular}[c]{@{}c@{}}Tightly-\\ coupled\\ (proposed)\end{tabular}}} \\ \cline{4-7} 
 &      &  & \textbf{N=1} & \textbf{N=2} & \textbf{N=3} & \textbf{N=4} \\ \hline
MH01 & 0.188            & 0.081    & 0.031 & 0.029 & 0.022 & 0.022  \\
MH02 & 0.140           & 0.085     & 0.036 & 0.032 & 0.027 & 0.025  \\
MH03 & 0.133             & 0.110   & 0.048 & 0.039 & 0.034 & 0.033  \\
MH04 & 0.186            & 0.119     & 0.068 & 0.058 & 0.051 & 0.048 \\
MH05 & 0.306            & 0.115    & 0.056 & 0.044 & 0.039 & 0.039  \\
V101 & 0.061            & 0.081    & 0.041 & 0.036 & 0.034 & 0.034  \\
V102 & 0.103             & 0.097   & 0.048 & 0.042 & 0.036 & 0.035  \\
V103 & 0.179            & 0.099    & 0.068 & 0.050 & 0.047 & 0.042  \\
V201 & 0.065              & 0.087   & 0.038 & 0.027 & 0.026 & 0.026 \\
V202 & 0.103              & 0.127  & 0.046 & 0.038 & 0.036 & 0.033  \\
V203 & 0.232            & 0.177    & 0.098 & 0.074 & 0.057 & 0.057  \\ \hline
\end{tabular}\label{table:rmse_euroc}
\end{table}
In Fig. \ref{fig:pose-error-V203}, we show the relative pose error, computed as proposed in~\cite{geiger2012we}, in the sequence V203, which is labeled as difficult. The top-view of the trajectory is in Fig \ref{fig:trajectory-V203}.
We see in the top plot of Fig. \ref{fig:pose-error-V203} that the relative translation error visibly decreases when global residuals are included in the estimator. 
Adding more than one error term per keyframe (i.e., N $>$ 1) helps improving the estimates. Increasing N from 1 to 2 reduces the ATE from 0.098 m to 0.074 m as shown in the bottom line of Table \ref{table:rmse_euroc}. %
The ATE decreases by 0.017 m with further increase of N to 3.
Increasing N from 3 to 4 does not provide any improvement.

The rotation error is also decreased by the addition of the global positional measurements in the sliding window optimization as shown in the bottom plot of Fig. \ref{fig:pose-error-V203}. However, the effect is smaller compared to the improvement achieved on the translation error. 
This result was expected since the estimator has only access to global positional information (i.e., no global orientation). 

\begin{figure}[t]
\begin{center}
\includegraphics[width=1.0\linewidth]{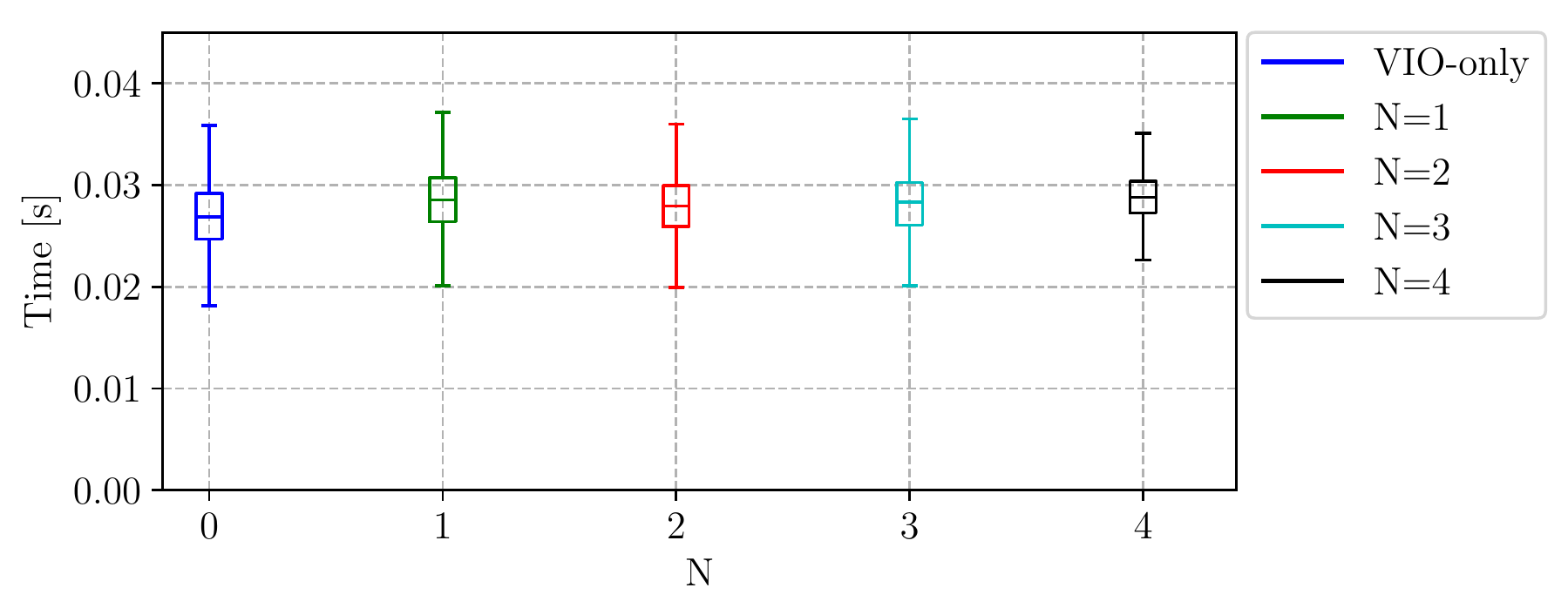}
\end{center}
   \caption{Optimizer time with respect to N. The value on the y axis represents the processing time, which is the duration between the time at which the front-end receives an image and the time at which its optimized pose is available from the sliding window optimization.}
\label{fig:timing}
\end{figure}

\textbf{Timing:} Increasing the value of N only slightly affects the processing time. The processing time was defined as the duration between the time at which the front-end receives an image and the time at which its optimized pose is available from the sliding window optimization. 
In Fig.~\ref{fig:timing}, we evaluate how the processing time varies with respect to N and compare to the VIO-only case.
The median is 26.2 ms for the VIO-only case and 27.7 ms for N = 1. 
The increase of just 1.3 ms shows that the formulation of the global residuals in (\ref{eq:global_position_residual}) allows to efficiently include new measurements in the sliding window optimization. As observed in Fig.~\ref{fig:timing}, adding more residual terms has a very negligible impact on the processing time.

\begin{figure}[t]
\begin{center}
\includegraphics[width=1.0\linewidth,height=1.0\linewidth]{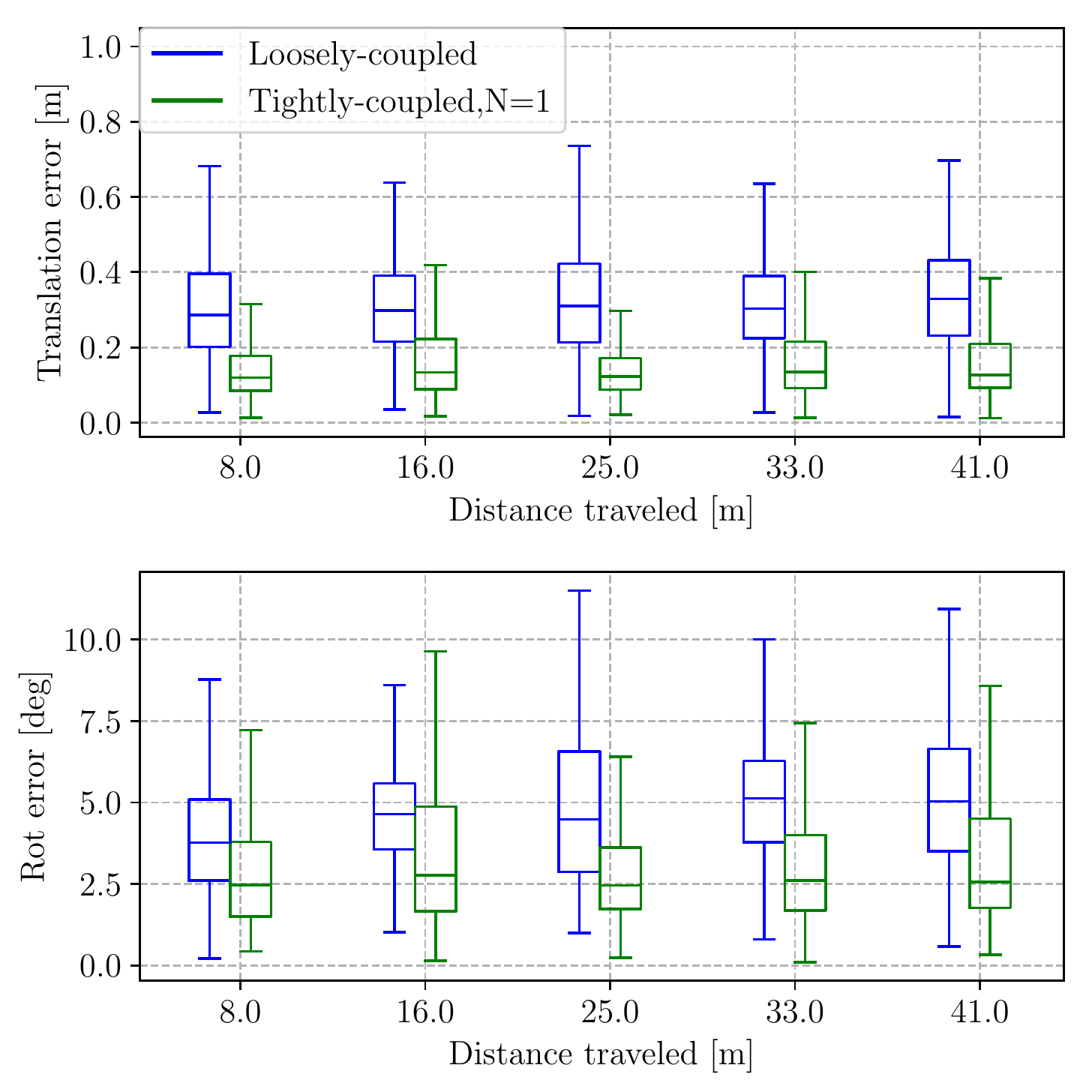}
\end{center}
   \caption{Relative translation and rotation error in EuRoC V203 difficult. Comparison between the tightly-coupled fusion approach proposed in this work and the loosely-coupled method based on \cite{qin2019general}. We used N=1 in the tightly-coupled method.}
\label{fig:pose-error-tightlt-vs-loosely-V203}
\end{figure}

\textbf{Comparison to loosely-coupled:} The proposed tightly-coupled fusion was compared to a loosely-coupled approach based on the method proposed in~\cite{qin2019general}. 
The loosely-coupled pose-graph optimization runs on a sliding window that contains the most recent keyframes selected by the VIO algorithm and the global position measurements. 
The newest frame in the sliding window corresponds to the most recent frame processed by the VIO pipeline.
Each keyframe is connected to one global measurement.
At every optimization step, the transformation between the VIO local frame and the global frame is estimated. %
Global position measurements are expressed with respect to such global frame. This transformation is applied to the most recent VIO output to obtain drift-free global pose estimates at the same rate of the VIO estimates.
We refer to~\cite{qin2019general} for more details on the loosely-coupled approach.
Fig. \ref{fig:pose-error-tightlt-vs-loosely-V203} shows the results of the two methods on sequence V203. Our tightly-coupled method outperforms the loosely-coupled approach in terms of both translation and rotation error with N $\in$ [1,2,3,4] in every EuRoC sequence, as shown in Table~\ref{table:rmse_euroc}.
Due to the noise in the global position measurements, the loosely-coupled fusion provides estimates less accurate than the VIO pipeline for sequence V202. 
\subsection{Outdoor Dataset with GPS Measurements}\label{UAVexperiment}
\subsubsection{Setup}
\begin{figure}[t]
\begin{center}
\includegraphics[width=1.0\linewidth]{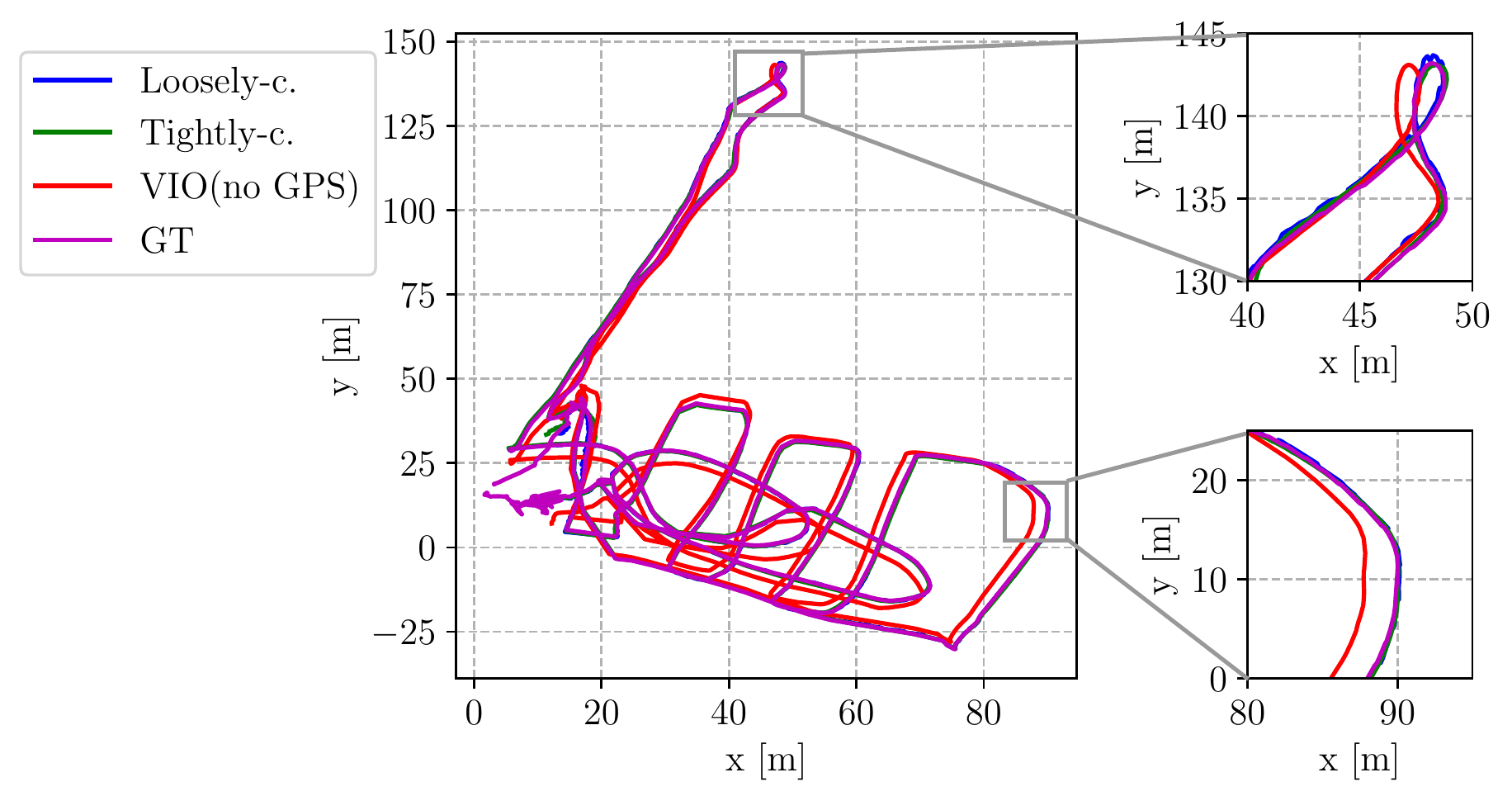}
\end{center}
   \caption{Top-view trajectory in flight sequence 3. Ground truth (GT), VIO (no GPS), loosely-coupled, and tightly-coupled trajectories are depicted.}
\label{fig:gomsf_trajectory}
\end{figure}

In this second set of experiments, we evaluated our approach on the dataset kindly provided by the authors of~\cite{mascaro2018gomsf}.
This dataset contains three flight sequences from an UAV equipped with a commercial stereo visual-inertial sensor, GPS, and ground-truth from a Leica total station.
The three flight sequences have a travelled distance of $404.1$ m, $483.3$ m, and $1033.3$ m, respectively.
The GPS data, acquired at 5 Hz, provides the global positional measurements. For our monocular visual front-end, we only used images from the left camera. 
Due to the unavailability of ground-truth orientation, we only reported the position error (mean and standard deviation) after the alignment.
Fig.~\ref{fig:gomsf_trajectory} shows the top-view of the third flight sequence, which contains a $1033.3$ m long trajectory.
For initialization, the initial position corresponds to the first GPS measurement.

\subsubsection{Results}
As observed in Table~\ref{table:ate_gomsf}, our tightly-coupled approach gives more accurate position estimates than the loosely-coupled.
The largest improvement is 50$\%$ in the second flight sequence and the smallest is 31$\%$ in the third flight sequence.

In the same table, we also included the best results of the loosely-coupled method proposed in~\cite{mascaro2018gomsf}, named GOMSF. GOMSF differs from~\cite{qin2019general} by the addition of a virtual node representing the local coordinate frame. VIO estimates are expressed in such coordinate frame. 
We can observe that our method also improves the mean position error with respect to GOMSF by 18$\%$, 14$\%$, and 12$\%$, respectively, in all three flight sequences. The difference in improvement with respect to~\cite{mascaro2018gomsf} is very likely due to the different front-end utilized: while we use the monocular SVO front-end, GOMFS uses the stereo OKVIS front-end~\cite{leutenegger2015keyframe}.
\begin{table}[]
\caption{Position error for the Outdoor Dataset}
\begin{tabular}{clcccc}
\hline
\textbf{Flight} &
  \textbf{\begin{tabular}[c]{@{}l@{}}Position\\ Error {[}m{]}\end{tabular}} &
  \textbf{\begin{tabular}[c]{@{}c@{}}VIO\\ (no GPS)\end{tabular}} &
  \textbf{\begin{tabular}[c]{@{}c@{}}Loosely-\\ coupled\\ (\cite{qin2019general})\end{tabular}} &
  \textbf{\begin{tabular}[c]{@{}c@{}}GOMSF\\ (\cite{mascaro2018gomsf})\end{tabular}} &
  \textbf{\begin{tabular}[c]{@{}c@{}}Tightly-\\ coupled\\ N=1\\ (proposed)\end{tabular}} \\ \hline
\multirow{2}{*}{1} & mean & 0.83 & 0.64 & 0.33 & \textbf{0.28} \\
                   & std  & 0.40 & 0.24 & 0.16 & \textbf{0.13} \\ \hline
\multirow{2}{*}{2} & mean & 1.28 & 0.35 & 0.29 & \textbf{0.24} \\
                   & std  & 0.63 & 0.17 & 0.13 & \textbf{0.09} \\ \hline
\multirow{2}{*}{3} & mean & 3.63  & 0.45 & 0.43 & \textbf{0.38} \\
                   & std  & 1.59 & 0.20 & 0.20 & \textbf{0.18} \\ \hline
\end{tabular}\label{table:ate_gomsf}
\end{table}
\section{Conclusion}\label{sec:conclusion}

Visual and inertial measurements are suitable to obtain locally accurate pose estimates but accumulate large drift in long-term navigation. To achieve high-rate, accurate, locally and globally consistent estimates, global positional information can be fused with visual and inertial measurements.
We proposed in this paper a tightly-coupled optimization-based methodology to solve the multi-sensors fusion problem.
We formulated the fusion problem as a keyframe-based sliding window optimization where the global measurements are employed to derive the new global factors. We leveraged the computation of the IMU preintegrated terms to include the global positional factors in the optimization with negligible increase of the computational cost compared to the visual-inertial case.
Experimental results showed that the proposed approach efficiently achieves accurate and globally consistent position estimates and consistently outperforms the state-of-the-art loosely-coupled approach.

{\small
\bibliographystyle{IEEEtran}
\bibliography{IEEEabrv,bibliography}
}

\end{document}